\documentclass[letterpaper]{article} 
\usepackage{aaai25}  
\usepackage{times}  
\usepackage{helvet}  
\usepackage{courier}  
\usepackage[hyphens]{url}  
\usepackage{graphicx} 
\usepackage{natbib}  
\usepackage{caption} 
\usepackage{amsmath}
\usepackage{multirow}
\usepackage{subfig}
\usepackage{enumitem}
\usepackage{soul}
\usepackage[ruled]{algorithm2e}
\usepackage{amssymb}
\usepackage{booktabs}

\title{Understanding Individual Agent Importance in Multi-Agent System \\via Counterfactual Reasoning}
\author{
    Jianming Chen\textsuperscript{\rm 1},
    Yawen Wang\textsuperscript{\rm 1}\thanks{Corresponding authors.},
    Junjie Wang\textsuperscript{\rm 1}\footnotemark[1],
    Xiaofei Xie\textsuperscript{\rm 2},
    Jun Hu\textsuperscript{\rm 1},\\
    Qing Wang\textsuperscript{\rm 1},
    Fanjiang Xu\textsuperscript{\rm 1}\footnotemark[1]
}
\affiliations {
    \textsuperscript{\rm 1}
    Institute of Software Chinese Academy of Sciences, Beijing, China
    \\
    Science \& Technology on Integrated Information System Laboratory, Beijing, China
    \\
    State Key Laboratory of Intelligent Game, Beijing, China
    \\
    University of Chinese Academy of Sciences, Beijing, China
    \\
    \textsuperscript{\rm 2}
    Singapore Management University, Singapore
    \\
    \{jianming2023, yawen2018, junjie\}@iscas.ac.cn,
    xfxie@smu.edu.sg,
    \{hujun, wq, fanjiang\}@iscas.ac.cn
}

\begin{document}

\maketitle

\begin{abstract}
Explaining multi-agent systems (MAS) is urgent as these systems become increasingly prevalent in various applications.
Previous work has provided explanations for the actions or states of agents, yet falls short in understanding the black-boxed agent's importance within a MAS and the overall team strategy.
To bridge this gap, we propose EMAI, a novel agent-level explanation approach that evaluates the individual agent's importance.
Inspired by counterfactual reasoning, a larger change in reward caused by the randomized action of agent indicates its higher importance.
We model it as a MARL problem to capture interactions across agents. 
Utilizing counterfactual reasoning, EMAI learns the masking agents to identify important agents.
Specifically, we define the optimization function to minimize the reward difference before and after action randomization and introduce sparsity constraints to encourage the exploration of more action randomization of agents during training.
The experimental results in seven multi-agent tasks demonstrate that EMAI achieves higher fidelity in explanations than baselines and provides more effective guidance in practical applications concerning understanding policies, launching attacks, and patching policies.
\end{abstract}


\section{Introduction}
\label{sec:intro}

Recent years have witnessed sensational advances in reinforcement learning (RL) across many prominent sequential decision-making problems. 
As these problems have grown in complexity, the field has transitioned from using primarily single-agent RL algorithms to multi-agent RL (MARL) algorithms, which are playing increasingly significant roles in various domains, e.g., unmanned aerial vehicles \cite{uav1,uav2,uav3}, industrial robots \cite{robots1,robots2,robots3}, camera network \cite{MATE,human-est}, and auto-driving \cite{driving1}. 
However, deep RL policies typically lack explainability, making them intrinsically difficult for humans to comprehend and trust. This issue is even more pronounced in multi-agent systems (MAS) due to the interactions and dependencies among agents. To broaden the adoption of RL-based applications in critical fields, there is a pressing need to enhance the transparency of RL agents through effective explanations.

Although some in-training explainable RL approaches (e.g., credit assignment) can simultaneously provide intrinsic explanations of the model when accomplishing tasks, they cannot work in black-box settings.
Prior work on post-training explanations for the black-box agent can be roughly divided into two categories.
The first category offers the observation-level explanation, i.e., revealing the regions of features within the observations that exert the most significant influence on the decisions of agent \cite{ob2,ob3,ob4}.
The second category delves into step-level explanations, aiming to identify the time-steps that are most or least pivotal to the agent's ultimate reward \cite{vb1-highlight, vb3, statemask}. 
Although previous research on post-training explanation shows great potential in helping users understand the behavior of the black-box agent, they cannot assess the importance of an agent at any specific state within the MAS. 

In MAS, the increase in the number of agents significantly contributes to the complexity of team strategies, and each agent plays its unique role and cooperates with others towards a common goal. 
Evaluating the importance of individuals in MAS helps reveal potential issues and vulnerabilities in collaboration, such as low-contributing agents (i.e., ``lazy'' agents) that limit system performance, or excessively high individual contributions that may indicate a lack of cooperation among agents \cite{mas-lazy}.
This can then potentially lead to a better training strategy for improving the overall performance of MAS.
Additionally, by identifying the important agents at each state, targeted and efficient interventions (e.g., launching attacks and patching policies) can be carried out more effectively \cite{edge,statemask}.

We propose EMAI, a novel agent-level \textit{\textbf{E}}xplanation approach for the \textit{\textbf{MA}}S which pinpoints the \textit{\textbf{I}}mportance of each individual agent at every time-step (i.e., state). 
In this paper, the agent required to be explained is black-boxed and called the target agent.
Motivated by counterfactual reasoning, which assesses the importance of a factor or decision to an outcome by envisioning scenarios contrary to reality, we define importance as the change in reward resulting from the random actions of target agents. The more important the agents, the greater the effect of their random actions on the reward.
Intuitively, one might attempt to achieve explanations by performing multiple random actions and observing the resulting changes in rewards. However, due to the space explosion problem \cite{mas-space} inherent in MAS and the necessity for multiple random transformations per interpretation, this approach would be highly inefficient.
To address this challenge, this work aims to learn the policy that guides the selection of specific agents for action randomization at each time step to more accurately and cost-effectively reveal the importance of the agents.

In this sense, we introduce the concept of masking agents, which learns a policy to mask unimportant target agents (i.e., make them take random actions). The importance of the target agent can be represented by its masking probability.
Since the importance of agents may be manifested through joint actions with other agents or delayed effects in subsequent time-steps \cite{dependency}, we model the learning of masking agents as a MARL problem, which decides whether or not to mask the target agents at each time-step, to capture these dependencies between agents and across time-steps.
Then, we utilize Centralized Training with Decentralized Execution (CTDE) paradigm \cite{ctde} to address the challenges of a holistic evaluation of each agent’s value and the exponential growth of joint action space with the number of agents in MAS \cite{cooperative, exp-space}.
To train the masking agents, we design the optimization objective to minimize the difference in performance before and after masking the target agents.
Besides, we design the sparsity constraint to encourage the exploration of masking as many target agents as possible during training.

We evaluate EMAI in seven popular multi-agent tasks and compare it with three commonly used and state-of-the-art baselines.
Results show that the explanations derived from EMAI have higher fidelity, with relative improvement ranging from $11\%$ to $118\%$ compared to baselines. 
Besides, based on the results of a manual evaluation, EMAI can help understand the policies by marking important agents in the visualization.
Model attackers can use EMAI to identify critical agents for attacking. The attacks guided by EMAI show the best performance, 
with the relative improvement ranging from $14\%$ to $289\%$ compared to baselines.
Finally, users can enhance the performance of MAS through patching critical agents identified by EMAI. Compared to baselines, the greatest improvements are achieved when guided by EMAI.

The contributions of this paper are as follows:
\begin{itemize}[leftmargin=*]
\item
A novel agent-level approach for explaining MAS by learning the importance of agents, which models the problem as MARL to learn the policy (masking agents) to randomize the actions of unimportant target agents. 

\item
Experimental evaluation on the fidelity of EMAI on seven multi-agent tasks with promising performance, outperforming three commonly used and state-of-the-art baselines.

\item
The demonstration of practical applications of this work, by evaluating the effectiveness of understanding policies, launching attacks, and patching policies with guidance from EMAI.
\end{itemize}

\section{Related Work}
\label{sec:related}

\textbf{RL explanation.}
Existing research on RL explanation primarily focuses on \textit{in-training} explanations and \textit{post-training} explanations.
(1) The in-training explainable RL models aim to design RL training algorithms that can simultaneously provide interpretable intermediate outcomes, enabling users to understand how the agent makes decisions and accomplishes tasks. Examples of such approaches include hierarchical RL \cite{HRL2,HRL3}, model approximation \cite{ma1, ma2}, and credit assignment \cite{ca1,ca2}.
Since the main goal of these approaches is to train better RL models, the provided interpretation ability is often a byproduct and tends to lack accuracy \cite{vb2-lazymdp}.
More importantly, this explanation is provided by the model itself. It cannot be used to explain a black-box target agent, in which case one can only query the actions taken by the agent under specific observations.
While post-training explanation approaches can provide an interpretation under black-box settings.

(2) The post-training explanation approaches involve explaining the decision-making process and strategies of the target agent after it has been trained. 
According to the perspective of explanation objectives, existing post-training explanation approaches can be mainly divided into two categories. 
The first category focuses on the observation-level explanation, which explains the regions of feature in the agent's observations that have the most significant impact on decisions, such as constructing saliency maps \cite{ob1, ob2, ob3} and learning strategy representations \cite{ma2, ob4}.
Regarding the second category of approaches, most of them provide step-level explanations to indicate the critical time-steps throughout the episode for achieving the final reward, e.g., the value function-based approaches \cite{vb1-highlight, vb2-lazymdp, vb3} and the approaches learning state-reward relationships \cite{statemask, edge, airs}. 
However, they typically cannot assess the importance of each agent per time-step, which is quite crucial for MAS.

\textbf{Counterfactual reasoning.} 
Counterfactual reasoning is a widely-used approach for explaining supervised learning models, e.g., explaining image classification models \cite{counter-image1, counter-image2, counter-image3}. These approaches involve perturbing the input and observing the impact on the classification outcome to reveal the reasons behind the specific predictions of models. 
Notably, Shapley value \cite{shapley1} is a concept related to, but distinct from, counterfactual reasoning. It considers all possible subsets of combinations that include the target participant. The focus is on marginal contribution, which refers to the incremental increase or decrease brought about by adding or removing a participant. Calculating the Shapley value is costly, particularly when dealing with a large number of agents \cite{shapley2}.
COMA \cite{COMA} connects the counterfactual reasoning and credit assignment in MARL. However, it falls under the category of in-training explainable RL mentioned earlier, which is not sufficiently accurate and cannot explain black-box MAS.

In the post-training RL explanations, there are also studies utilizing the counterfactual reasoning for the observation-level explanation \cite{ob1, ob2} and state-level explanation \cite{statemask}. 
This work shares a similar idea with those using counterfactual reasoning, yet our focus is on the agent-level explanation at every time-step in MAS, which is crucial yet has not been explored in prior research. 
The dependencies between agents and across time-steps in MAS make the explanation challenging.

\section{Approach}
\label{sec:approach}

\subsection{Problem Definition}
\label{sec:definition}
We consider a problem setting where a MARL joint policy $\pi = \{\pi_1, ..., \pi_n\}$ has been well trained for $n$ agents in the MAS. At each time-step $t$ in an episode, $i$-th agent obtains a local observation $o_{t, i}$ from the global state $s_t$ according to the observation function. The policy $\pi_i: o_i \xrightarrow{} a_i$ denotes the individual policy of $i$-th agent, which takes action $a_{t, i}$, depending on its local observations $o_{t, i}$. The joint action $a_t = \{a_{t, 1}, ..., a_{t, n}\}$ leads to the next state $s_{t+1}$ with the state transition probability $P(s_{t+1}|s_t, a_t)$.
Thereby, a global reward $r_t$ is obtained according to reward function $R(s_t, a_t, s_{t+1})$. 

Considering the variance in the importance of agents at different time-steps,
we aim to explain the MAS containing $n$ agents, by identifying the importance of target agents at each time-step $t$, i.e.,  $imp=\{imp^1_{t}, ..., imp^n_{t}\}$.
Our approach works under the black-box setting where only each agent's observation and corresponding action decision can be queried, which is more rational and practical, i.e., the value function and parameters of target agents are unavailable.

\begin{figure*}[t]
\centering
\subfloat[The workflow of EMAI. \label{fig:workflow}]{
		\includegraphics[width = 0.40\textwidth]{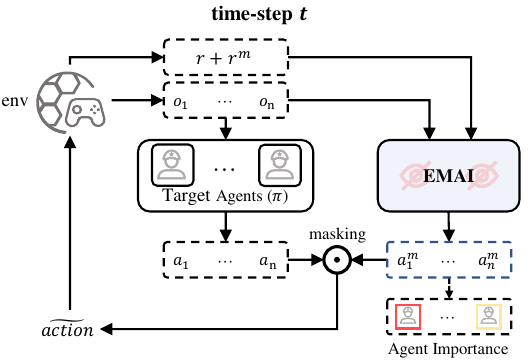}}
\subfloat[The architecture and training of EMAI. \label{training}]{
		\includegraphics[width = 0.58\textwidth]{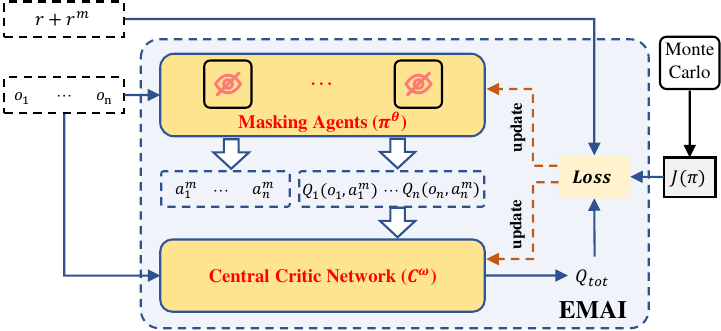}}
\caption{The overview of our proposed EMAI.
(a) At each time-step, EMAI outputs the masking probability for the action randomization of every target agent, and the lower probability indicates the higher importance of the corresponding target agent. (b) During the training, the masking agents' policy network learns the masking action and individual value, and the central critic network learns the total value to estimate the expected reward. The loss function is introduced to minimize the reward difference before and after the action randomization and encourage more action randomization of agents.}
\label{fig:framework}
\end{figure*}

\subsection{Problem Modeling}
\label{sec:modeling}
To measure the importance of a particular agent for the final reward, we draw inspiration from some works based on counterfactual reasoning \cite{counter-image3,statemask}. 
These approaches are based on the fundamental assumption that modifying the most important elements will exert the greatest impact on the outcome.
Similarly, in our problem, we can decide whether a particular agent is important or not by randomizing its actions at various time-steps and observing the change in the final reward. The importance of an agent can be reflected as ``how the final total rewards will change when its action is randomized''. If the reward difference is large, it indicates that the agent is highly important. Conversely, a minimal difference in rewards implies low importance.

We aim to learn a decision policy that generates a probability distribution, dictating the likelihood of selecting each agent for action randomization at each time step. The optimization goal of the policy is to minimize the reward difference before and after randomization. Thus, a lower probability of an agent being selected for randomization indicates its higher importance.
We model the learning of this decision policy as a MARL problem, to take into account the inter-agent and cross-time-step action dependencies (i.e., the cooperative relationships of joint actions and the delayed effects of actions across time-steps). 
Specifically, we introduce the EMAI, which incorporates our defined multiple masking agents.
For each $i$-th masking agent, we aim to develop a policy (denoted as $\pi_i^\theta$) that determines whether to randomize the action of the corresponding $i$-th target agent in the original MAS at each time-step, i.e., masking that target agent.

We treat the target agents with fixed joint policy $\pi$ as part of the environment.
Then the decision processes of masking agents is modeled as decentralized partially observable Markov decision processes (DEC-POMDP) \cite{DECPOMDP1, DECPOMDP2},
which can be defined by a tuple $G=< \mathcal{S}, \mathcal{A}^{m}, O, P, R, n, \gamma >$. 
The observation function $O$, state transition probability $P$, and reward function $R$ share the same definition in the origin environment of the target MAS.
$\mathcal{S}$ is the global state space.
$\mathcal{A}^{m}$ is our defined action space $\{0, 1\}$ of each masking agent.
The $a^{m}_{i \in \{1, \ldots, n\}} \in \mathcal{A}^{m}$ represents the masking action of the $i$-th masking agent, where $a^{m}_i=0$ denotes not masking and $a^{m}_i=1$ denotes masking the $i$-th target agent.
The observation $o_i$ for $i$-th agent is generated by the observation function $O(s,i)$. $\gamma$ is the discount factor applied to the rewards.
We define the policy of the $i$-th masking agent as $\pi^\theta_i: o_i \xrightarrow{} a^{m}_i$ parameterized by $\theta$.

Following the above, the workflow of our proposed EMAI is illustrated in Figure \ref{fig:workflow}. First, based on the observation generated from the state of the environment, each target agent takes actions through its fixed policy $\pi$. 
Meanwhile, EMAI trains the policy $\pi^\theta$ for masking agents to take masking actions at each time-step.
Then the final action for the $i$-th target agent $\tilde{a}_i$ is defined by the following operation:
\begin{equation}
    \label{equation:action_operation}
    \tilde{a_i} = a^{m}_i \odot a_i = \begin{cases}
        a_i,                    &if\ a^{m}_i=0,               \\
        random\ action,         &if\ a^{m}_i=1,
    \end{cases}
\end{equation}
where $a^{m}_i=0$ indicates that the $i$-th target agent retains its own action.
Otherwise when $a^{m}_i=1$, the action of target agent is replaced with a random action.
Then the final joint action $\widetilde{action} = \{\tilde{a_1}, ..., \tilde{a_n}\}$ that actually affects the environment can be acquired. 
Notably, we do not assume that EMAI only applies to discrete or continuous action space. 
In discrete action space, random action is randomly selected from a finite set $\{d_1, ..., d_k\}$ consisting of predefined $k$ discrete values.
If the action space is continuous, it is randomly sampled from an environmentally given continuous range $[lb, ub]$, e.g., to randomly select $0.88$ from the range $[-1, 1]$.

To ensure accurate masking of the low-importance target agent, we need a suitable objective function for training $\pi^\theta$.
The objective of $\pi^\theta$ is for masking the target agents while minimizing the following difference of the expected rewards:
\begin{equation}
    \label{equation:reward-diff}
    obj(\pi^{\theta}) = \arg\min_{\theta} \lvert J(\pi) - J(\pi^{\theta}) \rvert,
\end{equation}
where $J(\pi)$ represents the expected reward obtained by the target multi-agent with fixed policy $\pi$. 
Following previous research \cite{statemask}, $J(\pi)$ can be estimated as a constant in advance using the Monte Carlo method. Specifically, we have the target multi-agent run the game $500$ times and calculate the average expected discounted reward as follows: 
\begin{equation}
    \label{equation:est-exp-rew}
    J(\pi) = \mathbb{E}_{s_t, a_t}(\sum_t\gamma_t(R(s_t, a_t, s_{t+1})).
\end{equation}
The $J(\pi^\theta)$ in Equation \ref{equation:reward-diff} represents the expected reward when the actions of target agents are disrupted by the policy $\pi^\theta$ of masking agents, as follows:
\begin{equation}
    \label{equation:our-exp-rew}
    J(\pi^{\theta}) = \mathbb{E}_{s_t, a^{m}_t}(\sum_t\gamma_t(R(s_t, a^{m}_t, s_{t+1})).
\end{equation}

In addition, to encourage the exploration of masking more target agents during training, we set up a masking reward as sparsity constraint: $R^m(a^m_t) = \beta \sum^n_{i=1} a^m_{t, i}$, 
which reflects the number of target agents masked at time-step $t$. The $\beta$ is the weight hyper-parameter of the sparsity constraints.
The final total expected discounted reward is defined as:
\begin{equation}
    \label{equation:final-exp-rew}
    J^\prime(\pi^\theta) = J(\pi^{\theta}) + \mathbb{E}_{s_t, a^{m}_t}(R^m(a^{m}_t))).
\end{equation}

\subsection{The Architecture and Training of EMAI}
\label{sec:overview}
Due to the complex cooperation and diverse division of responsibilities within a MAS, the importance of each agent must be considered from a holistic perspective.
At the same time, the joint masking action space $\prod^n_{i=1}\mathcal{A}^{m}$ of all $n$ masking agents grows in an exponential manner with $n$.
Therefore, in our proposed EMAI, we apply CTDE \cite{ctde,DECPOMDP1} for MARL training.
In this framework, global information (including observations and actions of all agents) can be used to guide individual learning processes to consider their global impact, while each agent independently makes decisions based on its own observation, aiding in decomposing the joint action space.

As shown in Figure \ref{training}, EMAI consists of two networks:
the \(\pi^\theta\) of masking agents with weight parameters \(\theta\), and the central critic network \(C^\omega\) with weight parameters \(\omega\).
Specifically, as mentioned above, the network \(\pi^\theta\) learns the policy for the masking agents, based on the observations \([o_i]^n_{i=1}\), it outputs the masking actions $[a^m_i]^n_{i=1}$.
The network $C^\omega$ is constructed to evaluate the joint action of all masking agents from a global perspective.

Following value-based CTDE \cite{qmix,vdn}, we evaluate the value of masking actions when learning $\pi^\theta$.
Firstly, we train $\pi^\theta$ to learn individual value function $Q_i(o_i, a^m_i)$ for each one of the $n$ masking agents to assess individual policy, which represents the value of taking action $a^m_i$ for observation $o_i$.
Secondly, the $C^\omega$ learns the centralized value function $Q_{tot}(o, a^m)$ to evaluate the collective policy, which is the estimate of $J^\prime(\pi^\theta)$.

To ensure the maximization of both individual and total values simultaneously, it can be achieved by satisfying the following Individual-Global-Max (IGM) principle \cite{igm1,igm2}.
\begin{equation}
    \label{equation:igm}
    \begin{aligned}
    & \arg\max_{a^m\in \mathcal{A}^{m}} Q_{tot}(o, a^m; \omega) = \\
    & \left< \arg\max_{a^m_1} Q_1(o_1, a^m_1; \theta), ..., \arg\max_{a^m_n} Q_n(o_n, a^m_n; \theta) \right>.
    \end{aligned}
\end{equation}
Specifically, in the EMAI, we constrain the weights of the central critic network to be non-negative \cite{qmix} to ensure adherence to the Equation \ref{equation:igm},
which can be defined as the following form:
\begin{equation}
    \label{equation:fs}
    Q_{tot}(o, a^m; \omega) = \omega (Q_i(o_i, a^m_i; \theta)), \forall i \in \{1, ..., n\}; \omega \geq 0.
\end{equation}
The Equation \ref{equation:fs} can be replaced with other implements such as VDN \cite{vdn}, QPLEX \cite{qplex}, and QTRAN \cite{QTRAN}.

Finally, once $Q_{tot}$ has been determined, we use the following one-step TD loss \cite{DECPOMDP2} for the iterative optimization of $C^\omega$ and $\pi^\theta$, which minimizes the error between expected and estimate values $Q_{tot}(o, a^m)$.
\begin{equation}
    \label{equation:td-loss}
    \mathcal{L}_{e}(\omega, \theta) = \arg\min_{(\omega, \theta)} \mathbb{E} \left[ ((y_{tot} - Q_{tot}(o, a^m))^2\right],
\end{equation}
where $y_{tot} = reward + \gamma\max_{\hat{a}^m} \hat{Q}_{tot}(\hat{o}, \hat{a}^m)$, which denotes the expected value, and $reward$ is calculated by $R(s_t, a^{m}_t, s_{t+1}) + R^m(a^{m}_t)$, as introduced in Equation \ref{equation:final-exp-rew}.
The $\hat{Q}_{tot}(\hat{o}, \hat{a}^m)$ is 
calculated for the next time-step by the stale network. Previous researches \cite{targetnet1,na2q} have demonstrated the feasibility and stability of implementing updates in this manner.
Additionally, to minimize the difference between $J(\pi)$ and $J(\pi^\theta)$ as shown in Equation \ref{equation:reward-diff}, we calculate the following loss function:
\begin{equation}
    \label{equation:diff-loss}
    \begin{aligned}
    & \mathcal{L}_{d}(\omega, \theta) = \\
    & \arg\min_{(\omega, \theta)} \mathbb{E} \left[ (J(\pi) - \sum_t\gamma_t(Q_{tot}(o, a^m_t) - R^m(a^{m}_t)))^2 \right].
    \end{aligned}
\end{equation}
Thus, the total loss function for EMAI is:
\begin{equation}
    \label{equation:total-loss}
    \mathcal{L}_{total}(\omega, \theta) = \mathcal{L}_{e}(\omega, \theta) + \lambda\mathcal{L}_{d}(\omega, \theta),
\end{equation}
where $\lambda$ is the weighting term to balance the two loss functions.
Algorithm \ref{alg:train} briefly presents our training process.

\begin{algorithm}[tb]
	\caption{The training algorithm of EMAI.}
    \label{alg:train}
    \textbf{Input:} The policy $\pi$ of target agents, the original expected reward $J(\pi)$, the observations $\{o_1,...,o_n\}$    \\
    \textbf{Output:} The policy $\pi^\theta$ of masking agents    \\
    \textbf{Initialization:} The networks of $\pi^\theta$ and $C^\omega$    \\
    \For{each training batch}{
        Get original joint action from $\pi$: $\{a_1, ... , a_n\} = \pi(o_1,...,o_n)$  \\
        Get joint masking action and values from $\pi^\theta$: $\{a^{m}_1, ... , a^{m}_n\}, \{Q_1,...,Q_n\} = \pi^\theta(o_1,...,o_n)$  \\
        Get the final joint action: $\widetilde{action} = \{a_1, ... , a_n\} \odot \{a^{m}_1, ... , a^{m}_n\}$    \\
        Execute $\widetilde{action}$ of target agents and get $reward$ from environment   \\
        Calculate the global value $Q_{tot}$ using $C^\omega$ network    \\
        Update $\omega$ and $\theta$ by the TD loss with $reward$, $J(\pi)$, and $Q_{tot}$, as shown in Equation \ref{equation:total-loss}  \\
    }
	\label{alg1}
\end{algorithm}

\section{Experiments}
\label{sec:experiment}

We compare our proposed EMAI with three commonly-used and state-of-the-art RL explanation baselines in multiple widely-used multi-agent tasks. 
Following the existing studies \cite{statemask, edge}, we evaluate the fidelity of all explanation approaches and then validate the practicality of the explanations in terms of understanding policies, launching attacks, and patching policies.

\subsection{Experimental Setup}
\label{sec:exp_setup}

\subsubsection{Multi-Agent Environments}
Our experiments are conducted on three popular multi-agent benchmarks with different characteristics, selecting two to three environments from each benchmark as follows. 

\textbf{StarCraft Multi-Agent Challenge (SMAC).}
SMAC \cite{SMAC} simulates battle scenarios in which a team of controlled agents must destroy the built-in enemy team. SMAC is characterized by dense rewards and adversarial tasks. We consider three tasks in this environment which vary in the number and types of units controlled by agents.

\textbf{Google Research Football (GRF).}
GRF \cite{GRF} provides the scenarios of controlling a team of players to play football against the built-in team, characterized by sparse rewards and adversarial tasks. We choose two tasks in GRF, which vary in the number of players and the tactics.

\textbf{Multi-Agent Particle Environments (MPE).}
MPE \cite{MPE} consists of navigation tasks, where agents need to control particles to reach the target landmarks, which have the characteristics of dense rewards and cooperative tasks. We study two of these tasks, which mainly differ from whether explicit communication is required between the agents.

\subsubsection{Baseline Approaches}
We implement three popular and state-of-the-art baseline approaches in each multi-agent task to explain the importance of agents.
One approach works under the black-box setting, while the other two work under the white-box setting.

\textbf{StateMask} \cite{statemask}: the state-of-the-art post-training approach that analyzes the importance of the state for the final reward at each time-step.
When utilizing StateMask for interpreting the important agent in this paper, 
we treat the remaining agents as part of the environment and use it to compute the importance of time-step for each agent, thereby representing the importance of each agent at this time-step.

\textbf{Value-Based (VB)} \cite{vb3}:
a commonly-used in-training explanation approach for MAS, which represents the MARL work for the credit assignment or value decomposition problem and relates importance to the value function.
The value function is learned by addressing the credit assignment, e.g., the Q-value in QMIX \cite{qmix} or its variants \cite{qplex, na2q}.

\textbf{Gradient-Based Attribution (GBA)} \cite{GBA}: an approach for the in-training explanation that utilizes the gradients of the output logits, i.e., the log probability \( \log p(action_i) \), for explaining each agent's importance. 

\begin{figure*}[t]
    \centering
    \includegraphics[width=\textwidth]{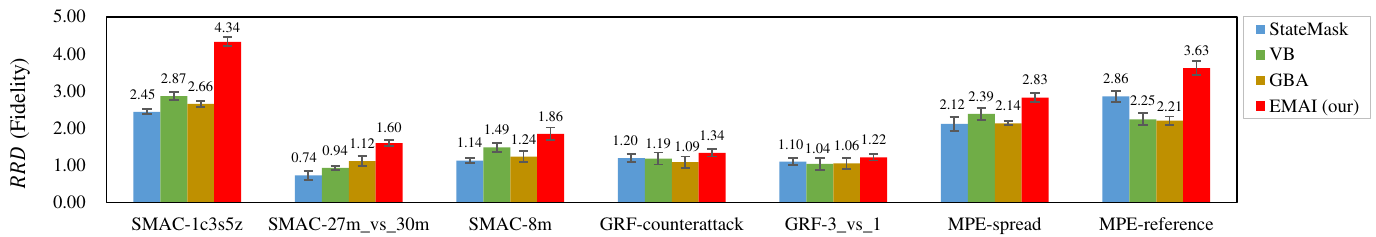}
    \caption{
    The results of the fidelity evaluation. The bar represents the mean value, and the black line on the bar represents the standard deviation.
    }
    \label{fig:rrd_res}
\end{figure*}

\begin{figure*}[t]
    \centering
    \subfloat[The critical agents in SMAC-27m\_vs\_30m. \label{fig:27m_vs_30m}]{
    		\includegraphics[width=0.99\textwidth]{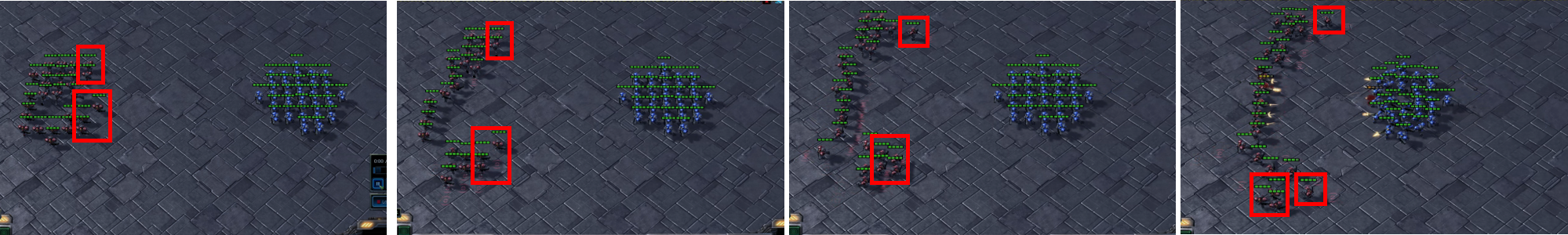}}
    \\
    \subfloat[The critical agent in GRF-counter\_attack. \label{fig:counter_attack}]{
    		\includegraphics[width=0.99\textwidth]{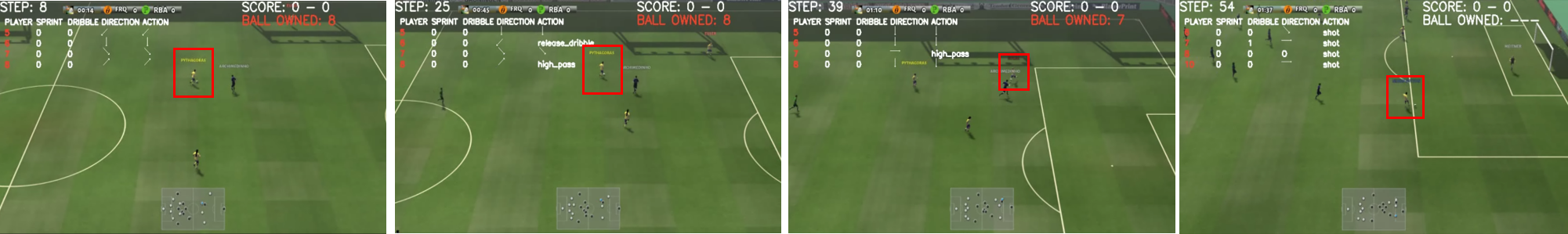}}
    \caption{
    The illustrations of EMAI identified critical agents, which is marked by the red box.}
    \label{fig:understanding}
\end{figure*}

\begin{table*}[t]
\centering
\small
\begin{tabular}{c|c|c|c|c}
\toprule
\textbf{Multi-Agent Tasks}      & \textbf{StateMask}       & \textbf{VB}      & \textbf{GBA}     & \textbf{EMAI} (ours)      \\
\midrule
SMAC-1c3s5z                     & -0.25 (0.08)             & -0.21 (0.16)     & -0.19 (0.09)     & \textbf{-0.68 (0.22)}        \\
SMAC-27m\_vs\_30m               & -0.48 (0.27)             & -0.35 (0.17)     & -0.45 (0.29)     & \textbf{-1.41 (0.18)}        \\
SMAC-8m                         & -0.46 (0.16)             & -0.53 (0.25)     & -0.41 (0.26)     & \textbf{-1.43 (0.33)}        \\
GRF-counter\_attack             & -3.45 (0.61)             & -3.19 (0.66)     & -3.16 (0.45)     & \textbf{-4.45 (0.31)}        \\
GRF-3vs1\_with\_keeper          & -1.89 (0.20)             & -1.63 (0.38)     & -1.34 (0.36)     & \textbf{-2.30 (0.28)}        \\
MPE-spread                      & -17.81 (8.49)            & -18.14 (7.78)    & -17.54 (6.60)    & \textbf{-23.57 (7.07)}       \\
MPE-reference                   & -5.40 (0.55)             & -6.24 (0.88)     & -5.09 (0.51)     & \textbf{-7.18 (0.92)}        \\
\bottomrule
\end{tabular}
\caption{The changes of episode team rewards before and after the attacks. The numbers outside and inside the parentheses represent the mean and standard deviation, respectively.}
\label{tab:attack_res}
\end{table*}

\begin{table*}[t]
\centering
\small
\begin{tabular}{c|c|c|c|c}
\toprule
\textbf{Tasks}            & \textbf{StateMask}      & \textbf{VB}         & \textbf{GBA}        & \textbf{EMAI (ours)}     \\
\midrule
SMAC-1c3s5z               & +0.19 (0.17)            & +0.37 (0.13)        & +0.22 (0.16)        & \textbf{+0.75 (0.24)}       \\
SMAC-27m\_vs\_30m         & +0.89 (0.16)            & +0.84 (0.18)        & +0.90 (0.22)        & \textbf{+1.11 (0.12)}       \\
SMAC-8m                   & +0.71 (0.41)            & +0.26 (0.51)        & +0.51 (0.59)        & \textbf{+0.92 (0.56)}       \\
GRF-counter\_attack       & +0.07 (0.64)            & -0.63 (0.64)        & +0.01 (0.55)        & \textbf{+1.44 (0.50)}       \\
GRF-3vs1\_with\_keeper    & +0.03 (0.42)            & -0.06 (0.58)        & -0.09 (0.46)        & \textbf{+0.33 (0.41)}       \\
MPE-spread                & +10.56 (1.39)           & +10.01 (0.92)       & +8.03 (1.24)        & \textbf{+12.57 (0.77)}      \\
MPE-reference             & +0.24 (1.04)            & +0.14 (0.80)        & +0.13 (1.06)        & \textbf{+0.72 (1.11)}       \\
\bottomrule
\end{tabular}
\caption{The changes of episode team rewards before and after the patching. The numbers outside and inside the parentheses represent the mean and standard deviation, respectively.}
\label{tab:patch_res}
\end{table*}

\subsection{Fidelity Evaluation}
\label{sec:fidelity}
\textbf{Evaluation Metric.}
Our work is able to identify the critical agents for obtaining the final reward at each time-step. Thus the fidelity of the explanation needs to measure the accuracy of the identified critical agents by demonstrating the high influence of these agents. 
Based on existing work \cite{statemask, edge, airs}, one intuitive way to assess fidelity is to randomize the actions of selected agents by the explanation approach and then measure the difference in reward before and after the action manipulation. If the selected agents are indeed critical to the final reward, then randomizing the actions of the critical agents should lead to greater reward variation compared to other agents.

Therefore, at each time-step, we select the most critical agent based on the explanation approach and randomize its action, while the rest of the agents act according to their policy decisions, thus resulting in the episode reward denoted as \(R_e\). 
The episode reward obtained by the agents' original actions is denoted as \(R_o\).
Additionally, the magnitude of reward variation may differ due to varying reward designs across different environments.
Therefore, based on the relative reward difference (\(RRD\)) introduced in previous work \cite{airs}, we use random selection to normalize the reward variation for each environment, i.e., we randomly select agents as critical ones to change its action, and the episode reward obtained is denoted by \(R_r\). Then the fidelity can be expressed as: $RRD= {\left|{R_e - R_o}\right|} / {\left|{R_r - R_o}\right|}$.

For each experiment, we perform 500 episodes and compute the mean value of episode rewards. The larger \(RRD\) represents better explanation fidelity. Note that if the reward variation is even smaller than the random selection (i.e., $RRD$ is less than $1$), then the explanation is extremely inaccurate.

\textbf{Result.}
Figure \ref{fig:rrd_res} compares the explanation fidelity of our proposed EMAI with baseline approaches in seven multi-agent tasks. It can be observed that EMAI achieves the highest RRD (i.e., fidelity) in all tasks, with the relative improvement ranging from $11\%$ to $118\%$ compared to baselines. This is because EMAI can accurately recognize the importance of each individual of multiple agents.
In contrast, StateMask's lack of effectiveness in the multi-agent setting is due to the fact that it focuses on the importance of a sequence of time-steps rather than on cross-sectional comparisons among agents.
VB also has lower fidelity scores compared to EMAI. This is because the value function is learned with the primary goal of guiding the agent to accomplish the task, while explaining the agent is merely a byproduct, leading to lower accuracy.
For similar reasons, GBA using log-probability fails to achieve good explanation performance.
In addition, in the task with the largest number of agents (SMAC-27m\_vs\_30m), the $RRD$ of the baseline approach is closer to or even less than $1$. It indicates that the critical agents selected by these explanation approaches are similar to, or even worse than, random selection.

Our proposed EMAI's superior performance compared to the baseline indicates its better interpreting ability. This advantage stems from the causal analysis abilities of counterfactual theory and the MARL approach's capacity to learn complex dependencies (both between agents and across time steps) in MAS. The combination effectively addresses the challenge of understanding agent importance in MAS.

\subsection{Practicability Evaluation}
Following existing work, we evaluate and analyze the practicality of the explanations provided by EMAI in three aspects: \textbf{understanding policies}, \textbf{launching attacks}, and \textbf{patching policies}. These reflect the practical significance of explanations for MASs.

\subsubsection{Understanding Policies}
\label{sec:understanding}
We visualize the critical agents identified by EMAI to demonstrate how EMAI helps humans understand the strategies of multi-agent. Due to space limitations, we present the results of two tasks: SMAC-27m\_vs\_30m and GRF-counter\_attack.


Figure \ref{fig:27m_vs_30m} illustrates a portion of time-steps in a winning episode of the target MAS (red team) in SMAC-27m\_vs\_30m. All agents are of the same unit type, yet the enemy holds a numerical advantage.
A key factor in accomplishing this task is the formation unfolding strategy, which is less intuitive and cannot be easily identified through unit hitting and being hit. EMAI successfully identifies the important agents on the flanks of the team.
Initially, these agents maneuver and spread out towards the upper and lower sides, establishing a semi-encirclement.
By maneuvering and dispersing their units, the red team can get greater firepower coverage and disperse the enemy's firepower to reduce the damage sustained by each unit. This strategy is pivotal in achieving the eventual victory.

Figure \ref{fig:counter_attack} displays a portion of time-steps in an episode of target MAS in GRF-counter\_attack. We use EMAI to identify the most critical agent in the yellow team. Initially, EMAI pinpoints the ball-carrying agent moving a long distance. Subsequently, after the ball is passed out for the first time, the agent identified by EMAI is the one running towards the ball's destination and passing it timely when the opponent's defense approaches. Finally, the agent recognized by EMAI is the one with a good shooting opportunity, receiving the ball, taking a shot, and successfully scoring. Therefore, EMAI provides insights into how each agent contributes to the final reward of the whole team. By focusing on these critical agents, we can naturally understand the team's goal-scoring strategy.

EMAI assigns an importance score to each agent at each time-step, facilitating a nuanced comprehension of the reasons underlying a MAS's ability or inability to fulfill the task.
Furthermore, we perform a user study to guarantee the objectivity of the evaluation. We mark the critical agents identified by various explanation methods in replays. Participants are invited to observe these replays and choose the explanation that corresponds to their intuition and effectively enhances the comprehension of policies. We enlist $36$ participants and equip them with the necessary background knowledge.
The survey results show that $75\%$ of participants believe that the explanations provided by EMAI are more aligned with human intuition,
and $58\%$ of participants believe that EMAI are helpful in identifying strategy flaws.
Therefore, EMAI is superior to all baseline approaches in understanding policies.

\subsubsection{Launching Attacks}
\label{sec:attack}

We experimentally analyze the significance of using the explanation approach to launch more effective attacks.
If the attack can be targeted towards the most critical agents, better effectiveness and covertness of attack may be easily achieved by only affecting these agents. 
Specifically, at each time-step, we target the most critical agent identified by the explanation approach and add
noise to its observations following common practice \cite{ob-attack}. Based on the perturbed observations, the agent may make suboptimal decisions according to its policy.

We conduct attack experiments for $500$ episodes, and the average changes of episode rewards before and after the attack are recorded in Table \ref{tab:attack_res}.
It can be observed that attacks guided by our proposed EMAI are the most effective (i.e., causing the most reduction in rewards), with the relative improvement ranging from $14\%$ to $289\%$ compared to the baselines.
This is due to the high-fidelity explanation for the agent importance provided by EMAI.

\subsubsection{Patching Policies}
\label{sec:patch}

We design a policy patching method guided by the explanation.
The core idea is to patch critical agents' actions to the one that is easily to gain a high reward.
In the process of explanation, we record the trajectories of critical agents' observations and corresponding actions in high-reward episodes to construct a patch package.
In the episode requiring patching, for the most critical agent identified by the explanation approach at each time-step, we search for an action corresponding to a similar observation in the patch package. 
The similarity between observations is calculated by Manhattan distance \cite{manhattan}. We search for the observation in the patch package that is close to the current observation. If the distance is below the threshold $d_{th}$, the observation is considered sufficiently similar. 
If multiple observations are found, the most similar observation is chosen.
Once a sufficiently similar observation can be found and the action output by the current policy is inconsistent with the actions in the patch package, we replace the policy-chosen action with the patch action.

We conduct experiments of patching for $500$ episodes, and Table \ref{tab:patch_res} shows the average changes of episode rewards before and after applying the patch to the critical agents. First, the patches guided by EMAI achieve the greatest improvement. Second, in some cases, several baselines could not guide patches correctly, even leading to a decrease in rewards.

\section{Conclusion}
\label{sec:conclusion}
This paper proposes EMAI, an approach for explaining MAS, which assesses the importance of individual agent based on counterfactual reasoning. Technically, we define masking agents to learn the importance evaluation of the target agents. The policy learning of masking agents is modeled as a MARL problem based on the CTDE paradigm, with the fundamental optimization objective to minimize the reward differences caused by counterfactual actions.
Experimental results show that compared to baselines, EMAI provides explanations with higher fidelity. Additionally, in three practical applications (i.e., understanding policies, launching attacks, and patching policies), EMAI also provides more effective guidance.

\textbf{Limitations and Future Works.}
First, we validated the practicality of using EMAI for attacking and patching target agents in some simple manners in the experiments. In the future, we will continue to explore how to leverage the explanations provided by EMAI to develop more effective attack and patch methods directed at important agents.
Additionally, EMAI identifies importance primarily based on the impact of agents' actions. In more complex environments, importance may also rely on other agent abilities, such as visual perception and planning abilities. We plan to investigate how to interpret policies in these more complex MAS scenarios, e.g., large swarms of drones.

\section{Acknowledgments}
This work was partially supported by the National Key Research and Development Program of China No.2021YFB3601400, National Natural Science Foundation of China Grant No.62232016 and No.62072442, Youth Innovation Promotion Association Chinese Academy of Sciences, Basic Research Program of ISCAS Grant No.ISCAS-JCZD-202304 and No.ISCAS-JCZD-202405, Major Program of ISCAS Grant No. ISCAS-ZD-202302, Innovation Team 2024 ISCAS (No. 2024-66), 
the National Research Foundation, Singapore, and the Cyber Security Agency under its National Cybersecurity R\&D Programme (NCRP25-P04-TAICeN). Any opinions, findings and conclusions or recommendations expressed in this material are those of the author(s) and do not reflect the views of National Research Foundation, Singapore and Cyber Security Agency of Singapore.

\small
\bibliography{aaai25}

\end{document}